\documentclass{article}
\usepackage{spconf,amsmath,graphicx}
\usepackage{comment}

\usepackage{comment}
\usepackage{multirow} 
\usepackage{color}
\usepackage{amssymb} 
\newcommand{\bvec}[1]{\mbox{\boldmath $#1$}}

\newcommand{\ts}{\textsuperscript} 
\newcommand{\bhline}[1]{\noalign{\hrule height #1}}
\usepackage{arydshln}            
\usepackage[subrefformat=parens]{subcaption} 

\usepackage{hyperref} 

\title{Efficient and Accurate Skeleton-based Two-person Interaction Recognition Using Inter- and Intra-body graphs}

\name{Yoshiki Ito$^{\dagger}$, Quan Kong, Kenichi Morita$^{\dagger}$, Tomoaki Yoshinaga$^{\dagger}$}
\address{R\&D Group, Hitachi, Ltd., Japan\\
$^{\dagger}$\normalsize{\texttt{\{yoshiki.ito.xf, kenichi.morita.ua, tomoaki.yoshinaga.xc\}@hitachi.com}}}

\begin{document}

\ninept
\maketitle




\begin{abstract}

\noindent
Skeleton-based two-person interaction recognition has been gaining increasing attention as advancements are made in pose estimation and graph convolutional networks.
Although the accuracy has been gradually improving, the increasing computational complexity makes it more impractical for a real-world environment.
There is still room for accuracy improvement as the conventional methods do not fully represent the relationship between inter-body joints.
In this paper, we propose a lightweight model for accurately recognizing two-person interactions.
In addition to the architecture, which incorporates middle fusion, we introduce a factorized convolution technique to reduce the weight parameters of the model.
We also introduce a network stream that accounts for relative distance changes between inter-body joints to improve accuracy.
Experiments using two large-scale datasets, NTU RGB+D 60 and 120, show that our method simultaneously achieved the highest accuracy and relatively low computational complexity compared with the conventional methods.

\end{abstract}



\section{INTRODUCTION}
\label{s:intro}

\vspace{-5mm}
\renewcommand{\thefootnote}{\fnsymbol{footnote}} 
\footnote[0]{\copyright 2022 IEEE. Personal use of this material is permitted. Permission from IEEE must be obtained for all other uses, in any current or future media, including reprinting/republishing this material for advertising or promotional purposes, creating new collective works, for resale or redistribution to servers or lists, or reuse of any copyrighted component of this work in other works.}
\renewcommand{\thefootnote}{\arabic{footnote}}

Human action recognition has been widely applied to many tasks in video understanding such as video surveillance, manufacturing, healthcare services, and human-computer interaction~\cite{pareek2021survey,majumder2020vision,zhang2019comprehensive}.
In particular, skeleton-based action recognition has been gaining increasing attention as the precision of human pose estimation has greatly \mbox{improved~\cite{geng2021bottom,yu2021lite}}.
Compared with approaches utilizing RGB images directly for action recognition, the skeleton-based approaches are more robust against changes in brightness, appearance, and interference from various background noise.

Recurrent neural network (RNN) and convolutional neural network (CNN) have often been utilized to recognize actions using skeleton data~\cite{liu2019skeleton,liu2017global,liu2017skeleton}. 
However, it is difficult for most of the models to recognize actions accurately due to the lack of the consideration of physical connections between the joints in the human body.
A graph convolutional network (GCN)~\cite{kipf2016semi} was introduced to the field of action recognition as a spatial-temporal GCN (ST-GCN)~\cite{yan2018spatial}, which was able to recognize actions more accurately by taking into account the structure of the body.
GCN-based methods have been explored over the past few years, and many extended methods based on ST-GCN have been proposed to date~\cite{feng2021multi,shi2020skeleton,shi2019two,li2019actional}.

\begin{figure}[t!]
  	\vspace{2mm}
  	\hspace{7mm}
  	\begin{minipage}[b]{0.4\linewidth}
    	\centering
    	\includegraphics[keepaspectratio, scale=0.16]{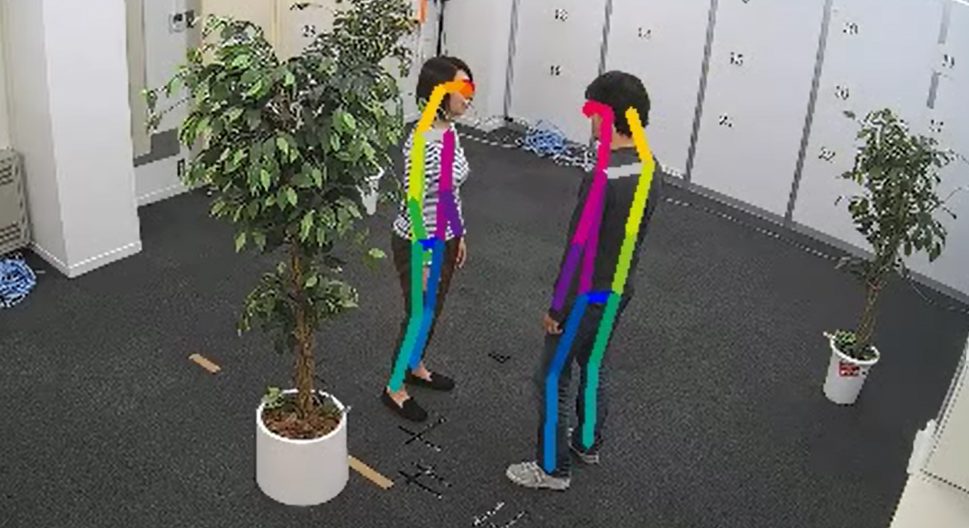}
    	\vspace{-1mm}
    	\subcaption{Talking}
  	\end{minipage}
  	\begin{minipage}[b]{0.38\linewidth}
    	\centering
    	\includegraphics[keepaspectratio, scale=0.163]{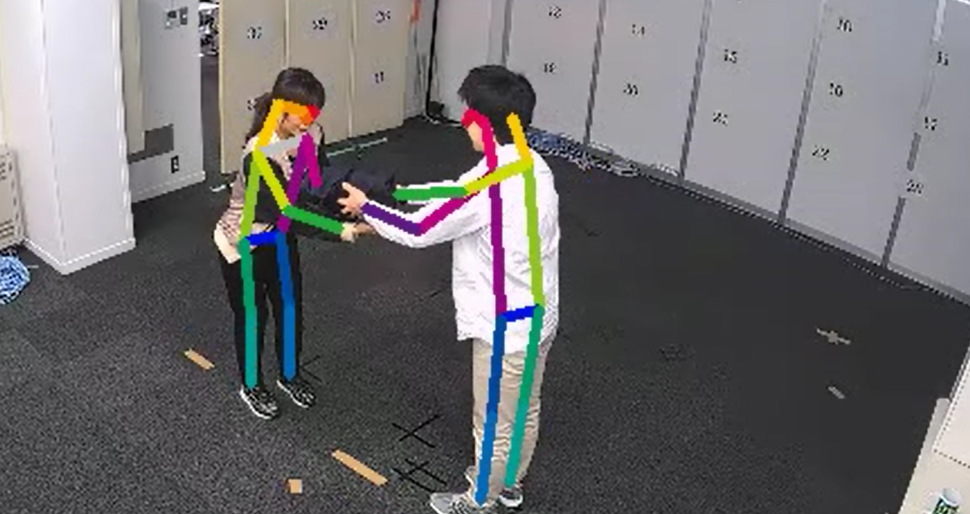}
    	\vspace{-1mm}
    	\subcaption{Giving object}
	\end{minipage}
	\vspace{-3mm}
	\caption{Examples of human interaction and their skeletons. Images were generated from MMAct dataset~\cite{kong2019mmact}.}
	\label{f:example}
	\vspace{-6mm}
\end{figure}

Two-person interaction recognition, which involves actions such as talking, hugging, and giving an object, as shown in Fig.~\ref{f:example}, is regarded as a more complex and challenging task than recognizing actions performed by an individual~\cite{zhang2019comprehensive,stergiou2019analyzing}.
This is because the relationship between two people needs to be represented for accurate interaction recognition.
As with individual action recognition, the GCN-based \mbox{methods~\cite{ito2021multi,zhu2021dyadic,yang2020pairwise}} are typically more accurate than the RNN or CNN-based \mbox{methods~\cite{perez2021interaction,nguyen2021geomnet,ye2020human}} because they have been designed to represent relationship between two bodies more accurately.

However, the conventional methods can still be improved in two aspects: computational complexity and recognition accuracy.
%
Although the accuracy is gradually improving with advancements in the GCN-based methods, the number of parameters has increased in most models.
For example, MS-AAGCN~\cite{shi2020skeleton} and MAGCN-IIG~\cite{ito2021multi} perform a heavy late fusion of scores calculated from four and eight network streams, respectively, to attain high accuracy.
It is more difficult to apply a model with a high computational cost in the real world because of its lower inference speed and need for larger computational resources.
Thus, it is necessary to design a lightweight architecture and introduce an effective computational method that does not degrade the accuracy.
%
The conventional methods do not consider the relationship between inter-body joints, which hinders their accuracy.
Since the methods do not sufficiently represent explicit inter-body relationship as inputs to a model, it is still difficult to distinguish two-person interactions.

\begin{figure*}[t!]
\centering
	\vspace{-1mm}
	\includegraphics[width=0.9\linewidth]{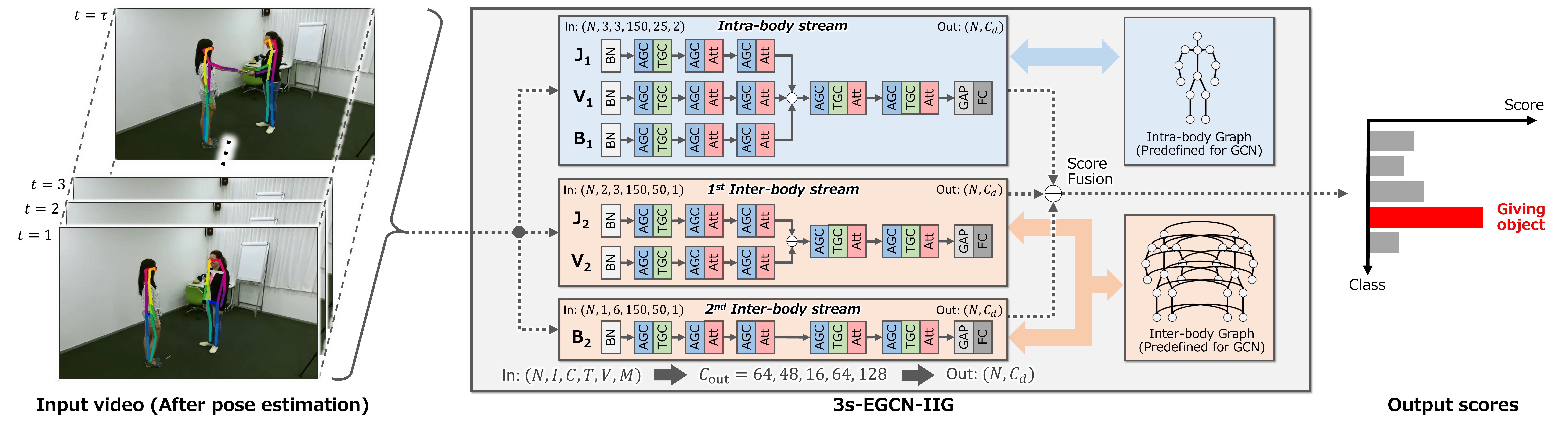}
	\vspace{-3mm}
	\caption{Overall architecture of 3s-EGCN-IIG (B0). The input shape to streams consists of six variables, $N,I,C,T,V,M$, which respectively denote batch size, number of branches, number of channels, temporal length, number of joints contained in the graph, and number of graphs used in the stream. $C_{\textrm{out}}$ and $C_d$ denote the number of output channels in each block and interaction classes. Note that the predefined graphs in this figure do not represent the actual connection, and the batch normalization and Swish function following each layer are omitted.}
	\label{f:flow}
	\vspace{-5mm}
\end{figure*}

To address the above problems, we aim to design an accurate model with fewer parameters.
Our model aggregates several streams as a few sets of branches to reduce the weight parameters.
The aggregated branches are fused in a stream in the form of middle fusion.
Additionally, to improve recognition accuracy, we propose a feature to capture inter-body relationships more accurately.
Our model accounts for the relative distance changes in inter-body joints using the novel features.
The contributions of this paper are summarized as follows.
\begin{itemize}
	\item The model complexity is greatly reduced by aggregating branches on the basis of input shape and graph type predefined for GCN.
	\item Recognition accuracy is improved by combining an inter-body graph-based stream capturing the change of relative distances with representative joints of the other person.
	\item Experiments using two large-scale human interaction datasets show that our method achieved the highest accuracy and low complexity compared with the state-of-the-art methods.
\end{itemize}


\vspace{-4mm}
\section{Proposed Method}
\label{s:method}
\vspace{-2mm}

To make the model lighter while improving the accuracy, our proposed method, three-stream efficient GCN using inter- and intra- body graphs (3s-EGCN-IIG), introduces three techniques for reducing computational complexity in \ref{ss:tech_complexity} and two techniques for improving accuracy in \ref{ss:tech_acc}.
The convolution and attention mechanism used in the method will be described in~\ref{ss:layers}.

\vspace{-4mm}
\subsection{Techniques for Reducing Computational Complexity}
\label{ss:tech_complexity}
\vspace{-2mm}

%
Several representative and effective techniques to reduce weight parameters have been introduced in EfficientGCN~\cite{song2021constructing} and other studies on individual action recognition task and object detection task.
The proposed method utilizes three techniques: (1) the aggregation of branches, (2) factorized convolution, and (3) the model scaling strategy.

\vspace{-4mm}
\subsubsection{Aggregation of Branches}
\label{sss:tech_complexity_aggr}
\vspace{-2mm}
The overall architecture of 3s-EGCN-IIG is shown in \mbox{Fig. \ref{f:flow}}.
It consists of three streams, the intra-body stream, the 1\ts{st} inter-body stream and the 2\ts{nd} inter-body stream.
Each stream has one or more branches to process joint coordinates (J$_x$), their velocity (V$_x$), or vectors between adjacent joints on the graph used in the stream (B$_x$).
Most of the conventional multi-stream models fuse several scores, which are outputs of streams, in the form of late fusion, but it results in many redundant parameters.
In contrast, our method uses a middle fusion approach~\cite{feng2021multi} to make the model lighter.
As shown in \mbox{Fig. \ref{f:flow}}, six branches are aggregated into three streams by the same input shape and graph type (e.g., intra-body graph and inter-body graph described in~\ref{sss:tech_acc_graphs}).
The branches are fused in the middle of each stream.
The final score is calculated as the average of the probability obtained from the three streams.

\vspace{-4mm}
\subsubsection{Factorized Convolution}
\label{sss:tech_complexity_fact}
\vspace{-2mm}

To further reduce the number of weight parameters, the convolutions are factorized into several steps.
An effective factorized convolution technique was originally introduced in MobileNetV2~\cite{sandler2018mobilenetv2} for object detection.
The technique largely reduces parameters by factorizing a convolution into two pointwise convolutions and one depthwise convolution.
The kernel size and stride settings are described in \ref{ss:layers}.

\vspace{-3mm}
\subsubsection{Model Scaling Strategy}
\label{sss:tech_complexity_scale}
\vspace{-2mm}
Several studies have investigated scaling strategies that efficiently increase the width and depth of the model.
An innovative strategy was proposed in EfficientNet~\cite{tan2019efficientnet} and applied to EfficientGCN.
The width and depth multipliers are $\alpha^\phi$ and $\beta^\phi$, respectively, under the constraint of $\alpha^2 \beta \simeq 2$ \mbox{$(\alpha, \beta \geq 1)$}, where $\phi$ is a compound coefficient.
The scaling strategy and its settings used in our method are set as $\alpha=1.2$ and $\beta=1.35$ in accordance with~\cite{song2021constructing}.
Our models are named \textit{3s-EGCN-IIG (B$\phi$)} depending on $\phi$.

\vspace{-4mm}
\subsection{Techniques for Improving Accuracy}
\label{ss:tech_acc}
\vspace{-2mm}

We introduce the following techniques for improving accuracy: (1) the collaborative use of the inter-body and intra-body graphs, and (2) an inter-body stream, which captures relative distances with the representative joints of the other person.

\vspace{-4mm}
\subsubsection{Intra-body Graph and Inter-body Graph}
\label{sss:tech_acc_graphs}
\vspace{-2mm}

Graphs that represent physical connections of the body must be predefined for graph convolution.
In this method, we utilize two graphs, the intra-body graph and the inter-body graph.
The intra-body graph was originally introduced in ST-GCN~\cite{yan2018spatial}.
In their method, an individual body structure is regarded as a graph, and then graph convolution is performed.
Since the intra-body graph is designed to focus only on each person's body, it is not appropriate to utilize it in the same way for a human interaction recognition task.

Consequently, a distinctive graph representation called the inter-body graph was proposed to capture the relationship between the joints of different bodies by means of graph convolutions~\cite{ito2021multi}.
The inter-body graph was defined by connecting the same kinds of joints between two bodies in addition to their intra-body connections as shown in \mbox{Fig. \ref{f:inter-body} (a)}.
The method collaboratively using the inter-body graph and the intra-body graph improved interaction recognition accuracy.
However, it is still difficult to sufficiently capture relationship between two bodies because the original inter-body graph only connects joints on the same side such as between left side of one person and the left side of the other person.
For instance, when passing an object, the opposite hands of two bodies may be used, e.g., one person's right hand and the other person's left hand.
If the graph connects only the same side, the joints on the opposite side are far apart on the graph and it may not properly take into account the important relationship between distant joints in two bodies.
Thus, we selected six representative joints in the human body (head, torso, and distal portion of the extremities, as shown in \mbox{Fig. \ref{f:inter-body} (b)}), and define all of their combinations as virtual edges.
In the example above, the novel inter-body graph takes into account the variety of which hand an object is given to, which should improve the accuracy.

\vspace{-4mm}
\subsubsection{Inputs into Branches}
\label{sss:tech_acc_input}
\vspace{-2mm}

The intra-body stream consists of three branches, J$_1$, V$_1$, and B$_1$, and utilizes the intra-body graph as its predefined graph.
The 1\ts{st} inter-body stream has J$_2$ and V$_2$ branches, which process joint coordinates and their velocity, respectively, using the inter-body graph.
In contrast, the 2\ts{nd} inter-body stream only has a B$_2$ branch and processes relative distances between joints of different bodies using the inter-body graph as well.

\begin{figure}[t!]
\centering
	\vspace{2mm}
	\includegraphics[width=0.79\linewidth]{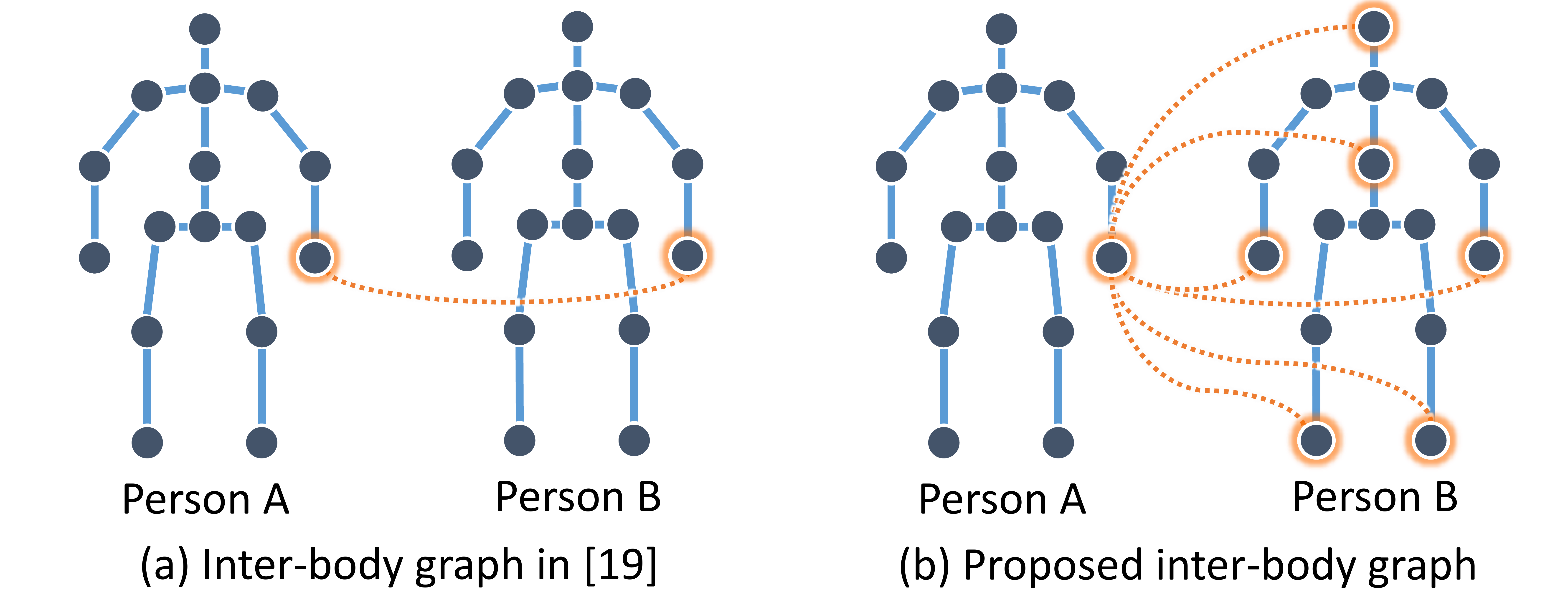}
	\vspace{-3mm}
	\caption{Illustration of joint connection on inter-body graph. Note that this figure shows only connections between a Person A's hand and Person B's corresponding joints. Relative distance features are also calculated using the six joints shown in (b).}
	\label{f:inter-body}
	\vspace{-5mm}
\end{figure}

Take 3D skeleton data with $x$, $y$, and $z$-coordinates as an \mbox{example}.
Given the joint $v_{p,i,t}=\{x_{p,i,t},y_{p,i,t,},z_{p,i,t}\}$, where $p \in \{p_1,p_2\}$ denotes a person index, $i$ denotes a joint index, and $t$ denotes a frame number, features input for the J$_1$ and J$_2$ branches are $f_{p,i,t}^{(\rm J_\cdot)} = v_{p,i,t}$.
The features are transformed into the input shape as shown in \mbox{Fig. \ref{f:flow}}.
Meanwhile, features input for the other branches are calculated as
\begin{align}
\label{e:gc}
	f_{p,i,t}^{(\rm V_\cdot)} &= v_{p,i,t}  - v_{p,i,t-1}, \\
	f_{p,ij,t}^{(\rm B_1)}    &= v_{p,i,t}  - v_{p,j,t}, \ j \in \mathcal{S}_i^\textrm{(intra)}, \\
	f_{ij,t}^{(\rm B_2)} &= \{\| v_{p,i,t} - v_{p',j_k,t} \|_2\}_{k=1,\cdots,6},\ j_k \in \mathcal{S}_i^\textrm{(inter)},
	%
	%
	\label{e:rdf}
\end{align}
where $\mathcal{S}_i^{(\cdot)}$ is defined as a joint set connected to the $i$-th joint on the graph.
$\mathcal{S}_i^\textrm{(intra)}$ includes only one adjacent joint, whereas $\mathcal{S}_i^\textrm{(inter)}$ includes six representative joints as with the definition of the inter-body graph as shown in \mbox{Fig. \ref{f:inter-body} (b)}.
In \mbox{Eq. (\ref{e:rdf})}, $p'$ is the person with whom $p$ interacts.
Particularly in the B$_2$ branch, relative distance features
are introduced by calculating distances between a person's $i$-th joint and the other person's six representative joints.
The input into the B$_2$ branch is given as 6-channel data by calculating the L2 norm and concatenating them.
%
%
Since the B$_2$ branch directly calculates relative distance features with all of the representative joints of the other person, it should improve accuracy for two-person interaction recognition.

\vspace{-3mm}
\subsection{Convolution and Attention Mechanism}
\label{ss:layers}
\vspace{-2mm}

The principal layers used in 3s-EGCN-IIG are adaptive graph convolution (AGC), temporal graph convolution (TGC), and attention (Att) as shown in Fig.~\ref{f:flow}.
The details of these layers will be described in this order.
Given $\bvec{f}_\textrm{in}(v_j)$ as the input feature vector on the $j$-th joint $v_j$ and $\mathcal{S}_i$ as a set of adjacent joints for the $i$-th joint $v_i$, the graph convolution operation on joint $v_i$ is written as
\begin{align}
\label{e:gc}
	\bvec{f}_\textrm{out}(v_i) = \sum_{v_j \in \mathcal{S}_i} \frac{1}{Z_{ij}} \
	\bvec{f}_\textrm{in}(v_j) \cdot w(l_i(v_j)).
\end{align}
A spatial partitioning strategy for distinguishing node characteristics is set in the same manner as that of the spatial configuration partitioning~\cite{yan2018spatial,shi2019two}; that is, the convolution kernel size is set to 3 and $\mathcal{S}_i$ is divided into three groups according to a distance between the root node and the center of gravity.
%
%
$l_i(v_j)$ denotes the label map at joint $v_j$ and is determined by the partitioning strategy.
$w(\cdot)$ denotes the function for obtaining the weight on the basis of the kernel index.
The normalizing term $Z_{ij}$ denotes the cardinality of the subsets to balance the contributions of different subsets.
Methods to optimize the topology of the predefined graphs and lower the dependence on the graphs have been proposed~\cite{shi2019two,sahbi2021learning}.
Our method utilizes AGC~\cite{shi2019two}.
%
\mbox{Eq. (\ref{e:gc})} is transformed as
\begin{align}
\label{e:agc}
	\bvec{f}_\textrm{s} = \sum_{k=1}^{K_v} \bvec{W}_k \bvec{f}_\textrm{in} \
	(\bvec{A}_k + \bvec{B}_k +\bvec{C}_k),
\end{align}
where $\bvec{W}_k$ denotes the weight matrix trainable through graph convolution and $K_v$ denotes the kernel size of the spatial dimension which is set to 3 by the spatial partitioning strategy.
$\bvec{A}_k$, $\bvec{B}_k$, and $\bvec{C}_k$ are the $V \times V$ matrices.
$\bvec{A}_k$ is a normalized adjacency matrix that represents physical connections in the human body, which is calculated as
$\bvec{A}_k = \bvec{\Lambda}_k^{-\frac{1}{2}} \overline{\bvec{A}}_k \bvec{\Lambda}_k^{-\frac{1}{2}}$ using the adjacency matrix $\overline{\bvec{A}}_k$.
$\overline{A}_k^{ij}$, the element $(i,j)$ of $\overline{\bvec{A}}_k$, is set to 1 or 0 depending on whether a joint $v_j$ is contained in the subset of joint $v_i$.
The diagonal elements of the normalized diagonal matrix $\bvec{\Lambda}_k$ are set as $\Lambda_k^{ii} = \sum_j(\overline{A}_k^{ij}) + \beta$, where $\beta$ is a small parameter to avoid empty rows.
$\bvec{B}_k$ is a trainable adjacency matrix that also represents the strength of the connections.
$\bvec{C}_k$ is a data-driven matrix based on the similarity of two joints.
The input feature map $\bvec{f}_\textrm{in}$, whose size is $(C_\textrm{in},T,V)$, is embedded by two embedding functions, $\zeta_k$ and $\eta_k$.
Each function is one $1 \times 1$ convolutional layer.
The output sizes of the functions are $(V,C'T)$ and $(C'T,V)$, and then they are multiplied to obtain a $V \times V$ similarity matrix.
The series of calculations to obtain $\bvec{C}_k$ from $\bvec{f}_\textrm{in}$ can be written as
$\bvec{C}_k = \textrm{softmax}(\bvec{f}_\textrm{in}^\top \bvec{W}_{\zeta_k}^\top \bvec{W}_{\eta_k} \bvec{f}_\textrm{in})$, where $\bvec{W}_{\zeta_k}$ and $\bvec{W}_{\eta_k}$ denote the weight of the above embedding functions.
The utilization of three matrices makes it possible to be less dependent on the predefined graphs.

TGC essentially follows AGC as shown in Fig.~\ref{f:flow}.
TGC can extract temporal context efficiently by using $\Gamma \times 1$ convolution, where $\Gamma$ denotes the temporal kernel size, which is set to 9, to control the temporal range.
The convolution can be formulated as
\begin{align}
\label{e:tgc}
	\begin{aligned}
		\bvec{f}_\textrm{t} (v_t) =& \
		\sum_{v_q \in \mathcal{S}(v_t)} \bvec{f}_\textrm{s} (v_q) \bvec{w}(v_q),\\
		\mathcal{S}(v_t) =& \
		\left\{ v_q \middle|\ |q-t| \leq \left\lfloor \Gamma/2 \right\rfloor \right\},
	\end{aligned}
\end{align}
where $v_t$ and $v_q$ denote the joints in the target frame $t$ and its neighbor frame $q$, respectively.
$\bvec{w}(v_q)$ denotes the weight for $v_q$ and $\mathcal{S}(v_t)$ denotes the sampling region of the temporal convolution.
As mentioned in~\ref{sss:tech_complexity_fact}, to reduce the number of weight parameters,
the convolution is factorized into three convolution steps: a $1 \times 1$ convolution from channel size $C_{\rm in}$ to $2C_{\rm in}$, a $5 \times 1$ depthwise convolution with a stride of 2, and a $1 \times 1$ pointwise convolution from channel size $2C_{\rm in}$ to $C_{\rm out}$.

An attention module called ST-JointAtt was also introduced in~\cite{song2021constructing}.
ST-JointAtt was inspired by~\cite{hou2021coordinate} and designed to distinguish the most informative joints in certain frames from the whole sequence by taking spatial and temporal information into account concurrently.
The formulation is written as 
\begin{align}
\label{e:att}
	\begin{aligned}
		\bvec{g}_\textrm{inner} = \
		&\theta (p_s(\bvec{g}_\textrm{in}) \oplus p_t(\bvec{g}_\textrm{in})) \cdot \bvec{W}),\\
		\bvec{g}_\textrm{out} = \
		&\bvec{g}_\textrm{in} \odot (\sigma(\bvec{g}_\textrm{inner} \cdot \bvec{W}_s) \
		\otimes \sigma(\bvec{g}_\textrm{inner} \cdot \bvec{W}_t)),
	\end{aligned}
\end{align}
where $\bvec{g}_\textrm{in}$ and $\bvec{g}_\textrm{out}$ denote input and output feature maps.
$p_s(\cdot)$ and $p_t(\cdot)$ represent spatial and temporal average pooling operations, respectively.
The symbols $\oplus$, $\odot$, and $\otimes$ respectively denote the concatenation, the channel-wise outer-product, and the element-wise product.
$\theta(\cdot)$ and $\sigma(\cdot)$ respectively denote HardSwith and Sigmoid activation function.
$\bvec{W} \in \mathbb{R}^{C \times \frac{C}{4}}$ and $\bvec{W}_s, \bvec{W}_t \in \mathbb{R}^{\frac{C}{4} \times C}$ are trainable weight matrices.
The concatenation of two vectors obtained from each pooling operation makes it possible to calculate attention maps using both spatial and temporal information.


\begin{table*}[t]
	\begin{center}
	\caption{Comparison of accuracy (\%), number of parameters, and FLOPs between proposed method and state-of-the-art methods on 11 interactions in NTU RGB+D 60 and 26 interactions in NTU RGB+D 120. Results that were not provided are marked with ``-''.}
	\vspace{-2mm}
	\label{t:result}
	\scalebox{0.90}[0.85]{
  	\begin{tabular}{l l l : c c c c : c c}
  		\bhline{1.3pt}
	    \multirow{2}{*}{Methods} & \multirow{2}{*}{Conf./Jour.} & \multirow{2}{*}{Type} & \multicolumn{2}{c}{NTU RGB+D 60} & \multicolumn{2}{c:}{NTU RGB+D 120} & \#Param. & FLOPs \\
	    &&& X-Sub & X-View & X-Sub & X-Set & ($\times$ M) & ($\times$ G) \\ \hline
	    ST-LSTM~\cite{liu2017skeleton}    & TPAMI (2017) & RNN & 83.0 & 87.3 & 63.0 & 66.6 & - & -\\
	    GCA-LSTM~\cite{liu2017global}     & CVPR (2017)  & RNN & 85.9 & 89.0 & 70.6 & 73.7 & - & -\\
	    ST-GCN~\cite{yan2018spatial}      & AAAI (2018)  & GCN & 91.0 & 94.7 & 82.5 & 83.6 & 3.10&16.3\\
	    FSNET~\cite{liu2019skeleton}      & TPAMI (2019) & CNN & 74.0 & 80.5 & 61.2 & 69.7 & - & -\\
	    AS-GCN~\cite{li2019actional}    & CVPR (2019) & GCN & 89.3 & 93.0 & 82.9 & 83.7 & 9.50&26.8\\
	    2s-AGCN~\cite{shi2019two}       & CVPR (2019) & GCN & 92.4 & 95.8 & 86.1 & 88.1 & 6.94&37.3\\
	    ST-GCN-PAM~\cite{yang2020pairwise}& ICIP (2020) & GCN & -    & -    & 82.4 & 88.4 & - & -\\
	    LSTM-IRN~\cite{perez2021interaction} & TMM (2021) & RNN & 90.5 & 93.5 & 77.7 & 79.6 & - & -\\
	    2s-DRAGCN~\cite{zhu2021dyadic}    & Patt. Recog. (2021) & GCN & 94.7 & 97.2 & 90.6 & 90.4 & 7.14 & - \\
	    MAGCN-IIG~\cite{ito2021multi}   & Access (2021) & GCN & 94.4 & 97.5 & 89.0 & 93.1 & 29.6&143\\
	    GeomNet~\cite{nguyen2021geomnet}  & ICCV (2021) & CNN & 93.6 & 96.3 & 86.5 & 87.6 & -&-\\
	    EfficientGCN (B0)~\cite{song2021constructing} & TPAMI (2022) & GCN & 95.0 & 97.1 & 89.6 & 90.8 & 0.39 & 2.08\\
	    \hline
	    3s-EGCN-IIG (B0) & \multirow{2}{*}{ICIP (2022) *Ours} & \multirow{2}{*}{GCN} &\bf{96.3}&\bf{98.3}&\bf{91.9}&\bf{94.8}&0.96&4.65\\
	    3s-EGCN-IIG (B4) &        &     &\bf{96.6}&\bf{98.7}&\bf{92.4}&\bf{95.5}&6.01&22.5\\
	    \bhline{1.3pt}
    \end{tabular}
    }
    \end{center}
    \vspace{-6mm}
\end{table*}





\vspace{-2mm}
\section{EXPERIMENTS}
\label{s:exp}
\vspace{-2mm}

\vspace{-2mm}
\subsection{Datasets}
\label{ss:datasets}
\vspace{-2mm}

We evaluated our proposed method through experiments on two large-scale human action datasets, NTU RGB+D 60~\cite{shahroudy2016ntu} and NTU RGB+D 120~\cite{liu2019ntu}, which contain 11 and 26 interactions, respectively.
These datasets have been widely utilized for evaluating action recognition methods due to their large data size.
%
Each class contains approximately 1,000 clips.
The skeleton data was obtained using Kinect v2 sensors and represented by 3D coordinates for 25 joints per person.
These datasets include many variations in subjects and camera set-ups.
The original papers recommend two kinds of validation methods to split the data into a training set and a test set in terms of subjects and camera set-ups, i.e., cross-subject (X-Sub) and cross-view (X-View).
Note that X-View is called cross-setup (X-Set) in NTU RGB+D 120.

\vspace{-2mm}
\subsection{Training Details}
\label{ss:training_details}
\vspace{-2mm}

The training processes for both datasets are the same, as follows.
They included 50 epochs in total.
A warmup strategy~\cite{he2016deep} was applied over the first 10 epochs to gradually increase the learning rate from 0.0 to 0.1 for stable training.
After the 10\ts{th} epoch, the learning rate decays according to a cosine schedule~\cite{loshchilov2016sgdr}.
The stochastic gradient descent with Nesterov Momentum
was applied as the optimization strategy and a hyperparameter was set to 0.9.
Cross-entropy was selected as the loss function and the weight decay was set to 0.0001.
The batch size was set to 32. 
Since each clip contains multiple frames with different lengths, the length of all clips was aligned to 150 frames (5.0 seconds at 30 fps) for the input into the streams.
%
%
%
The code was implemented by PyTorch 1.7.0.

\vspace{-2mm}
\subsection{Results and Discussion}
\label{ss:results}
\vspace{-2mm}

We compared our proposed method with the state-of-the-art methods, the results of which are shown in Table~\ref{t:result}.
Note that we implemented ST-GCN, 2s-AGCN, and MAGCN-IIG ourselves and evaluated their accuracy.
The other scores were taken from the original papers of the respective methods.
As shown in the table, our method showed the highest accuracy in all four validations, even with the smallest B0 model, achieving an average improvement of 2.3\% in NTU RGB+D 120 compared with the conventional highest performing model (MAGCN-IIG).
It should also be noted that the number of parameters of the proposed B0 model is still less than 1M.
The middle fusion in each stream achieved high accuracy while reducing the number of layers efficiently.
A middle fusion-based network had roughly half with the number of parameters of a late fusion-based network which requires more layers.
The factorized convolutions further reduced the number of parameters by approximately 10\%.
Although the B4 model is larger than B0, the accuracy of the B4 model is on average 0.6\% higher than that of the B0 model in NTU RGB+D 120, and the computational complexity is still lower than that of most of the conventional methods.
We measured the inference speed using 1 NVIDIA GeForce RTX 3090 GPU.
The average inference speed of B0 model was approximately 5 ms per sample.
%

Ablation studies were also conducted to evaluate the performance and effectiveness of each stream and the combined streams.
The comparisons are shown in Table~\ref{t:ablation}.
The methods which combined two or more streams yielded higher accuracy than the single stream methods.
In particular, 3s-EGCN-IIG combining three streams achieved the highest classification accuracy.
This is because the inter-body streams are largely conducive to more accurate classification, with an average improvement of 3.2\% although the two streams only have 0.57M parameters.
%
%
Therefore, it is important to utilize the inter-body graph and consider the relative distances from the joints of the other person to express human interaction.

\begin{table}[t]
	\begin{center}
	\vspace{1mm}
	\caption{Comparison of accuracy (\%), number of parameters, and GFLOPs with different input streams using the proposed method in NTU RGB+D 120. All models are based on B0.}
	\label{t:ablation}
	\vspace{-2mm}
	\scalebox{0.90}[0.85]{
  	\begin{tabular}{l c c c c}
  		\bhline{1.3pt}
	    Methods                        & X-Sub     & X-Set     & \#Param. & FLOPs \\ \hline
	    (A) Intra-body stream          & 89.6      & 90.8      & 0.39 M   & 2.08 G \\
	    (B) 1\ts{st} inter-body stream & 89.6      & 91.5      & 0.32 M   & 1.50 G \\
	    (C) 2\ts{nd} inter-body stream & 86.7      & 91.5      & 0.25 M   & 1.07 G \\ \hline
	    (A)+(B)                        & 91.6      & 93.4      & 0.71 M   & 3.58 G \\
	    (A)+(C)                        & 90.8      & 94.0      & 0.64 M   & 3.15 G \\
	    (B)+(C)                        & 90.3      & 94.0      & 0.57 M   & 2.57 G \\ \hline
	    (A)+(B)+(C)                    & \bf{91.9} & \bf{94.8} & 0.96 M   & 4.65 G \\
	    \bhline{1.3pt}
    \end{tabular}
    }
    \end{center}
    \vspace{-8mm}
\end{table}


\vspace{-2mm}
\section{CONCLUSION}
\label{s:concl}
\vspace{-2mm}

We have proposed 3s-EGCN-IIG for efficient and accurate human interaction recognition.
The aggregation of branches, factorized convolution, and the scaling strategy were utilized to greatly reduce the computational complexity so that the method can be applied in real-world environments.
Our method utilized both the inter-body graph and intra-body graph in different streams for accurate human interaction recognition.
The relative distance features, which are calculated on the basis of distances with the representative joints of the other person, were used in one of the inter-body streams, which contributed to improvements in accuracy.
Through the experiments using two large-scale datasets containing a wide variety of human interactions, we showed that our method successfully reduced the number of weight parameters while achieving high accuracy.


\vfill
\pagebreak

\end{document}